\documentclass[conference]{IEEEtran}

\IEEEoverridecommandlockouts

\usepackage{cite}
\usepackage{amsmath,amssymb,amsfonts}
\usepackage{algorithmic}
\usepackage{graphicx}
\usepackage{booktabs}
\usepackage{subcaption}
\usepackage{textcomp}
\usepackage{xcolor}
\usepackage{makecell}
\usepackage{url}
\def\BibTeX{{\rm B\kern-.05em{\sc i\kern-.025em b}\kern-.08em
    T\kern-.1667em\lower.7ex\hbox{E}\kern-.125emX}}

\begin{document}

\title{Breath-Dx: A Time Series Classification Benchmark for Disease Screening with eNose
\\[0.45em]{\large\bfseries IEEE ICDM 2026 --- Best Paper Runner-Up (Applied Track)}}

\author{
\IEEEauthorblockN{1\textsuperscript{st} Ivan Poddiakov}
\IEEEauthorblockA{\textit{Sber AI Lab} \\
Moscow, Russia \\
ivanpodd@gmail.com}
\hspace{5cm}
\and
\IEEEauthorblockN{2\textsuperscript{nd} Ivan Kruzhilov}
\IEEEauthorblockA{\textit{Moscow Power Engineering Institute} \\
Moscow, Russia \\
kruzhilovis@mpei.ru}
\hspace{6cm}
\and
\IEEEauthorblockN{3\textsuperscript{rd} Galina Zubkova}
\IEEEauthorblockA{\textit{Sber AI Lab} \\
Moscow, Russia \\
galina.v.zubkova@gmail.com}
\hspace{5cm}
\and
\IEEEauthorblockN{4\textsuperscript{th} Svetlana Erofeeva}
\IEEEauthorblockA{\textit{Moscow Regional Research and} \\
\textit{Clinical Institute (MONIKI)}\\
Moscow, Russia \\
erofsb@mail.ru}
\hspace{5cm}
\and
\IEEEauthorblockN{5\textsuperscript{th} Andrey Savchenko}
\IEEEauthorblockA{\textit{Sber AI Lab}\\
\textit{Laboratory for Theoretical Foundations}\\
\textit{of AI Models, HSE University} \\
Moscow, Russia \\
avsavchenko@hse.ru}
\hspace{5cm}
\and
\IEEEauthorblockN{6\textsuperscript{th} Pavel Blinov}
\IEEEauthorblockA{\textit{SQB AI} \\
Tashkent, Uzbekistan \\
blinoff.pavel@gmail.com}
}


\maketitle

\begin{abstract}
Electronic nose (eNose) technology offers a portable, non-invasive route for disease screening via exhaled breath analysis. However, progress in the field is hindered by the lack of systematic benchmarks that compare machine learning methods under realistic conditions, particularly sensor drift, a known but rarely quantified failure mode. Most existing studies rely on small, private datasets with inconsistent protocols, rendering results non-comparable and slowing progress. We introduce Breath-Dx, the first comprehensive benchmark of time series classification methods for breath-based disease screening. Our work leverages the Scent of Health (S-OH) dataset, the largest public clinical eNose collection to date, comprising 1,234 patients across nine diagnostic groups, each providing a 17-channel multivariate time series with explicit temporal train/test splits that account for sensor drift. We systematically evaluate 10 methods spanning four families: feature-based tabular learners, convolution-based models, graph-based, and time series architectures. All methods degrade significantly under temporal validation (0.05-0.12 AUC drop compared to random cross-validation), confirming that drift embeds a strong temporal signal. Under this realistic evaluation, the best methods achieve ROC AUC of 0.74-0.76 for some lung conditions, though others remain near chance. Recurrent architectures show the most consistent performance across diseases, while specialized time series methods achieve the highest peak results. Demographic confounders matter: breath signals provide independent diagnostic value for some conditions but add little beyond age for others. We release the S-OH dataset, all temporal splits, evaluation protocols, and code to enable reproducible research and accelerate progress in clinical breathomics.
\end{abstract}


\section{Introduction}
\textbf{Motivation.} Recent advances in artificial intelligence have been largely driven by large-scale datasets and architectures that exploit spatial and temporal structures in data, such as convolutional networks and transformers~\cite{li2024survey}. However, despite impressive successes in vision and language, certain sensory modalities remain underserved by open data and benchmarks. One of the most notable is chemical sensing for breath analysis.

Exhaled breath contains complex mixtures of volatile organic compounds (VOCs) that encode clinically relevant metabolic information. Electronic nose (eNose) technology offers a portable, cost-effective route to digitize these signals for non-invasive diagnostics~\cite{lee2024cross}. eNose systems have demonstrated clinical promise for conditions including asthma, lung cancer, and tuberculosis~\cite{li2023research}. Yet transitioning prototypes into reliable clinical tools requires solving fundamental data science challenges: learning from high-dimensional time series, generalizing across sensor drift, and validating on diverse patient populations.

Despite growing interest, the field lacks systematic benchmarks that compare machine learning methods under realistic conditions. Existing studies typically evaluate a single method on small, private datasets (often with less than 100 samples) using inconsistent evaluation protocols~\cite{rodriguez2019ultrafast,tenero2020electronic,acevedo2021electronic}. Consequently, basic questions regarding which classifier families are most robust to sensor drift, or how performance scales with training data, remain unanswered. This opacity hinders progress and makes it difficult for practitioners to select appropriate methods for clinical deployment of eNose technology.

\textbf{Breath-Dx benchmark.} In this paper we introduce the Scent of Health (S-OH) dataset, the largest public clinical eNose collection to date. It comprises 1,234 patients across nine diagnostic groups, each providing a 17-channel multivariate time series of approximately 15 minutes (895 seconds). The dataset incorporates samples from two distinct clinical sites, making it the first public eNose resource to support cross-site generalization analysis. Critically, S-OH includes explicit temporal train/test splits that separate data by collection week, enabling controlled evaluation of sensor drift--a known but rarely quantified failure mode in chemical sensing~\cite{bosch2022electronic}. The dataset is released publicly to support reproducible benchmarking.

We leverage S-OH to construct Breath-Dx, the first comprehensive benchmark of time series classification methods for breath-based disease screening. We evaluate a diverse portfolio of 10 methods spanning four families of ML methods: (i) feature-based tabular learners (CatBoost, XGBoost, HistGradientBoost, RandomForest) with engineered kinetic signal features; (ii) deep learning models adapted from computer vision (CNN, ResNet18); (iii) graph-based (Graphormer); (iv) time series-specialized architectures (LSTM, InceptionTime, TS2Vec). All methods are evaluated under identical temporal splits that simulate realistic deployment conditions with sensor drift.

The main \textbf{contributions} of this work are:
\begin{itemize}
    \item We introduce the Breath-Dx benchmark, which provides the first reproducible quantification of sensor drift impact on clinical eNose data, revealing a 0.05-0.12 AUC drop under temporal validation and demonstrating that random cross-validation is inappropriate for such studies.

    \item We present the first systematic comparison of 10 time series methods on clinical breath data, showing that no single method dominates. LSTM is the most stable across diseases (top-3 for 5/9 conditions). InceptionTime and Graphormer achieve the highest peak performance (AUC$>$0.74) on specific diseases (lung cancer and COPD), but they lack cross-disease stability.

    \item We quantify actionable insights for clinical deployment: breath signals provide independent diagnostic value beyond demographics for liver disease and gastritis (sensor-based AUC$>$0.65 vs demographics-only AUC$\approx$0.5), while for lung cancer and tuberculosis, age alone rivals or exceeds sensor-based models (AUC=0.678 and 0.803, respectively), demonstrating that demographic confounders must be explicitly accounted for when evaluating eNose-based screening.
\end{itemize}

\section{Related Work}
\subsection{Clinical eNose Studies for Disease Detection}
Recent advances in sensor technology have enabled large-scale studies of exhaled breath analysis for disease diagnosis~\cite{goikhman2022quantum}. Prior clinical work has focused on specific organ systems, including the digestive tract~\cite{tiele2019breath}, lungs and respiratory tract~\cite{baldini2020electronic}, and the excretory system~\cite{capuano2025disposable}. However, these studies share three critical limitations that our benchmark directly addresses.

\textbf{Limitation 1: Small, single-pathology cohorts.} Current eNose studies in healthcare typically target individual diseases or anatomically-related conditions~\cite{mortazavi2025exhaled} and rely on very limited datasets of tens to hundreds of patients. Representative examples include chronic lung conditions (n=56, Accuracy 82\%)~\cite{rodriguez2019ultrafast}, lung cancer (n=145, Specificity 84\%)~\cite{van2018training}, asthma (n=38, AUC 80\%)~\cite{tenero2020electronic}, and tuberculosis (n=224, Specificity 87\%)~\cite{bruins2013diagnosis}. While these results are promising, the small cohort sizes raise concerns about statistical power and generalizability. Moreover, focusing on a single pathology per study prevents the development of models that can discriminate among multiple conditions--a realistic requirement in clinical screening settings.

\textbf{Limitation 2: No reproducible benchmarks.} In most clinical eNose studies, neither the raw sensor data nor the evaluation code are publicly released~\cite{leopold2015comparison}. This lack of transparency makes it impossible to compare methods across studies, reproduce reported results, or build upon prior work. As a result, progress in the field remains fragmented, with each research group effectively starting from scratch.

\textbf{Limitation 3: Inconsistent evaluation protocols.} Even when performance metrics are reported, evaluation protocols vary widely across studies. Some use random cross-validation, others use a simple train/test split, and very few account for temporal dependencies such as sensor drift~\cite{worner2025long,jaeschke2020enose}. This heterogeneity makes it difficult to assess whether improvements are due to better methods or more optimistic evaluation schemes.

\subsection{Public Olfactory Datasets: Molecular vs Clinical}
Beyond clinical studies, several public datasets have been released for olfactory research. However, they focus predominantly on molecular odor perception rather than disease screening. Examples include OlfactionBase~\cite{sharma2022olfactionbase}, M2OR~\cite{lalis2024m2or}, SmellNet~\cite{feng2025smellnet}, and SMILES Odors~\cite{lee2023principal}, which provide chemical structure-to-odor mappings for computational chemistry. Others, such as the Drift in MOX Gas Sensors dataset~\cite{worner2025long}, characterize sensor drift behavior but lack any clinical annotations. To the best of our knowledge, only one prior public eNose dataset exists for disease detection--a chronic lung conditions dataset comprising just 78 patients~\cite{acevedo2021electronic}. This dataset is limited to a single pathology and does not provide temporal splits to evaluate drift robustness.

In summary, the field lacks a large-scale, multi-pathology, multi-site clinical eNose dataset, a reproducible benchmark with standardized temporal splits that account for sensor drift, and a systematic comparison of diverse time series classification methods on a common platform. Our work fills this gap by introducing the Breath-Dx benchmark, providing the first large-scale public resource for clinical breathomics with explicit drift-aware evaluation protocols.

\section{Problem Formulation}
\subsection{Multivariate Time Series Classification for Breath Analysis}
Let a breath sample be represented as a multivariate time series:
\begin{equation}
X = \{x_1, x_2,..., x_T\} \in R^{T \times C}, \label{eq}
\end{equation}
where $T = 358$ is the number of time steps (895 seconds at 0.4 Hz sampling rate) and $C = 17$ is the number of sensor channels. Each $x_t \in R^{C}$ is the vector of sensor readings at time $t$.

Each sample $X$ is associated with a binary label $y \in \{0,1\}$ indicating the presence ($y = 1$) or absence ($y = 0$) of a target disease. The goal is to learn a classifier $f:R^{T \times C} \xrightarrow{} [0,1]$ that estimates the probability of disease given the breath signal.

We formulate this as a binary classification problem for each of the nine diagnostic groups separately (healthy controls plus eight diseases). For each disease, we adopt a one-vs-rest strategy where samples from the target disease form the positive class, while all other samples (including healthy controls and other diseases) form the negative class. This mirrors realistic screening scenarios where a clinician asks whether a patient has a specific condition, irrespective of other possible diagnoses.

\begin{table*}
\centering
\normalsize
\caption{Distribution and demographic characteristics of the S-OH dataset by diagnostic group.} \label{tab:cohort}
\begin{tabular}{lccccc}
\toprule
\textbf{Diagnostic Group (ICD-10 code)} & \textbf{N} & \textbf{\%} & \textbf{Age (mean)} & \textbf{Age (std)} & \textbf{Male (\%)} \\
\midrule
Healthy individuals (Z00) & 256 & 20.7 & 40.6 & 12.9 & 21.5 \\
Diabetes mellitus type II (E11) & 128 & 10.4 & 60.4 & 11.9 & 30.5 \\
Gastritis and duodenitis (K29) & 138 & 11.2 & 52.3 & 16.3 & 35.5 \\
Non-alcoholic fatty liver disease (K76) & 128 & 10.4 & 48.0 & 16.2 & 39 \\
Hepatitis B/C (B18) & 138 & 11.2 & 50.9 & 13.4 & 55 \\
Malignant neoplasm of lungs (C34) & 100 & 8.1 & 66.7 & 8.6 & 73 \\
Chronic renal failure (N18) & 128 & 10.4 & 59.8 & 14.6 & 50 \\
COPD (J44) & 100 & 8.1 & 64.2 & 10.4 & 64 \\
Respiratory tuberculosis (A15) & 118 & 9.5 & 41.9 & 16.1 & 62.7 \\
\midrule
\textbf{Total or Average} & \textbf{1,234} & \textbf{100} & \textbf{53.9} & \textbf{13.4} & \textbf{44.1} \\
\bottomrule
\end{tabular}
\end{table*}

\subsection{Temporal Splits: Simulating Sensor Drift} \label{Temporal_Splits}
A key challenge in real-world eNose deployment is \textit{sensor drift} -- gradual changes in sensor responses over time due to aging, contamination, or environmental factors~\cite{bosch2022electronic}. Standard random cross-validation artificially masks this problem by mixing temporally close samples across train and test sets, leading to overly optimistic performance estimates.

To evaluate robustness to drift, we adopt temporal validation~\cite{leopold2015comparison} by partitioning the collection weeks into training and testing sets such that: (i) all samples from a given patient belong entirely to either the training set or the test set (no patient leakage); (ii) training and test weeks are disjoint, ensuring that no temporal leakage occurs between training and evaluation; and (iii) for each disease, we select different temporal segments for evaluation, ensuring that test samples are temporally distant from training samples.

This design explicitly forces the classifier to generalize across time, revealing how performance degrades under realistic deployment conditions.

\subsection{Evaluation Metric and Linear Baseline}
We report performance using the Area Under the Receiver Operating Characteristic Curve (ROC AUC)~\cite{fawcett2006introduction} as it is well-suited for our task. First, our dataset is imbalanced, and AUC, unlike accuracy, is insensitive to class priors. Second, clinical screening often requires trading off sensitivity and specificity, and AUC summarizes performance across all possible operating points without imposing a fixed threshold. Third, AUC is the standard metric in breath analysis studies~\cite{tenero2020electronic, van2018training}. For each disease and each method, we report the mean and standard deviation of AUC over 16 random training splits (with a fixed temporal test set), using 15\% of the training data for validation and early stopping.

As a simple baseline, we train a Logistic Regression classifier with L2 regularization (default settings) on the concatenated raw signals from all channels. Each multivariate time series is flattened into a single vector, normalized using StandardScaler (zero mean, unit variance), and then fed into the classifier, which is trained for up to 1,000 iterations. This baseline is intended to provide a lower bound for comparison, helping to assess whether more sophisticated methods offer meaningful improvement over a linear model.

\section{S-OH Dataset Overview}
\subsection{Data Collection Protocol}
The S-OH dataset was collected over 13 consecutive weeks across two clinical sites: 11 weeks at the Moscow Regional Research and Clinical Institute (MONIKI) covering all disease groups except tuberculosis, followed by 2 weeks at the Central Research Institute of Tuberculosis (CRIT) enrolling only tuberculosis patients (A15) and healthy controls. The study protocol was approved by local ethics committees, and all participants provided written informed consent. Before sampling, participants adhered to a standardized protocol (4-hour fasting, no smoking for at least 4 hours, no alcohol for 48 hours, avoidance of perfumes). Each participant provided a single 2-liter exhaled breath sample collected in a sterile disposable bag and analyzed within 4 hours of collection.

\subsection{Hardware Overview}
The eNose device uses a multi-electrode chip with 17 co-planar electrode pairs, each bridged by a functional metal oxide material~\cite{fedorov2020microplotter}. The chip includes on-chip thermoresistors and heaters to maintain a stable operating temperature of $300 \pm 5^\circ C$. The functional materials (ZnO and metal-doped ZnO: In-ZnO, Ag-ZnO, Ce-ZnO, Ni-ZnO) were deposited via printing and annealed to remove residual solvent. The complete sensor array is enclosed in a gas-tight chamber with volume $0.76~cm^3$. The sampling rate is 0.4 Hz, producing 358 time steps per 895 seconds across all 17 channels.

\subsection{Cohort Composition}
Table~\ref{tab:cohort} summarizes the cohort composition. A total of 1,234 patients were enrolled, spanning nine diagnostic groups based on ICD-10 codes~\cite{ICD10}: healthy controls (Z00) and eight diseases: diabetes mellitus type II (E11), gastritis and duodenitis (K29), non-alcoholic fatty liver disease (K76), hepatitis B/C (B18), malignant neoplasm of lungs (C34), chronic renal failure (N18), chronic obstructive pulmonary disease (COPD) (J44), and respiratory tuberculosis (A15). These conditions were selected based on two criteria: (i) they represent leading causes of morbidity and mortality worldwide~\cite{gbd2019diseases,sung2021global}, and (ii) each has a known or suspected metabolic fingerprint in exhaled breath, as reported in prior clinical studies~\cite{li2023research,capuano2025disposable,baldini2020electronic}. The set spans multiple organ systems (liver, gastric, renal, pulmonary, metabolic), enabling evaluation of eNose technology across diverse pathological mechanisms.

\subsection{Temporal Structure and Drift-Aware Evaluation}
\begin{figure*}[tb]
    \centering
    \begin{subfigure}[b]{0.4\textwidth}
        \centering
        \includegraphics[width=\textwidth]{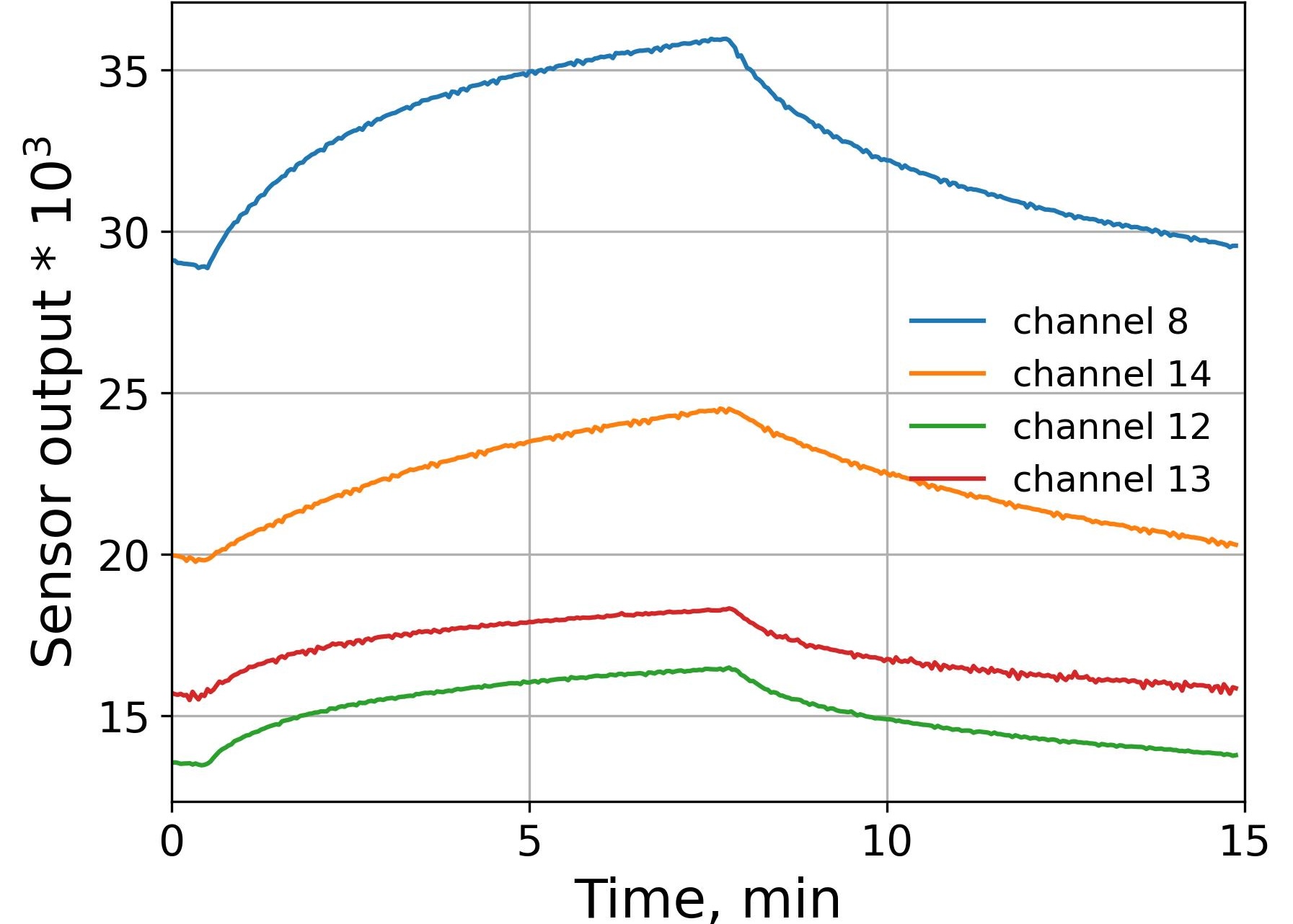}
        \caption{After 2 days of operation.}
        \label{fig:sub1}
    \end{subfigure}
    \quad
    \begin{subfigure}[b]{0.4\textwidth}
        \centering
        \includegraphics[width=\textwidth]{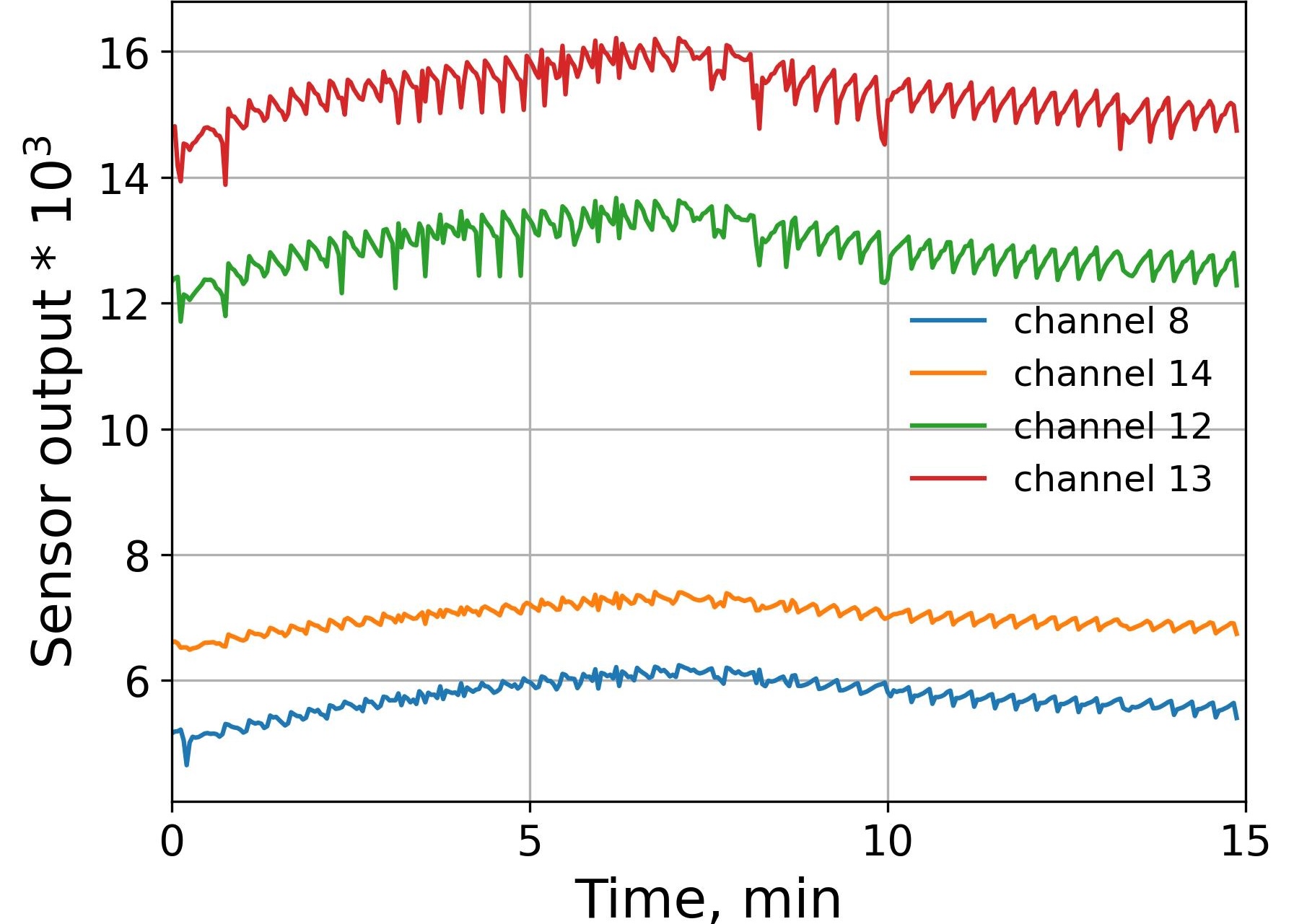}
        \caption{After 53 days of operation.}
        \label{fig:sub2}
    \end{subfigure}
    \caption{Progressive sensor signal deterioration.} \label{fig:drift}
\end{figure*}

Figure~\ref{fig:drift} illustrates the signal deterioration, showing waveforms from four representative channels (of the 17 total) at day 2 and day 53. The comparison reveals increased fluctuations and shifts in absolute signal levels over time.

\begin{figure}[t]
    \begin{center}
    \includegraphics[width=\linewidth]{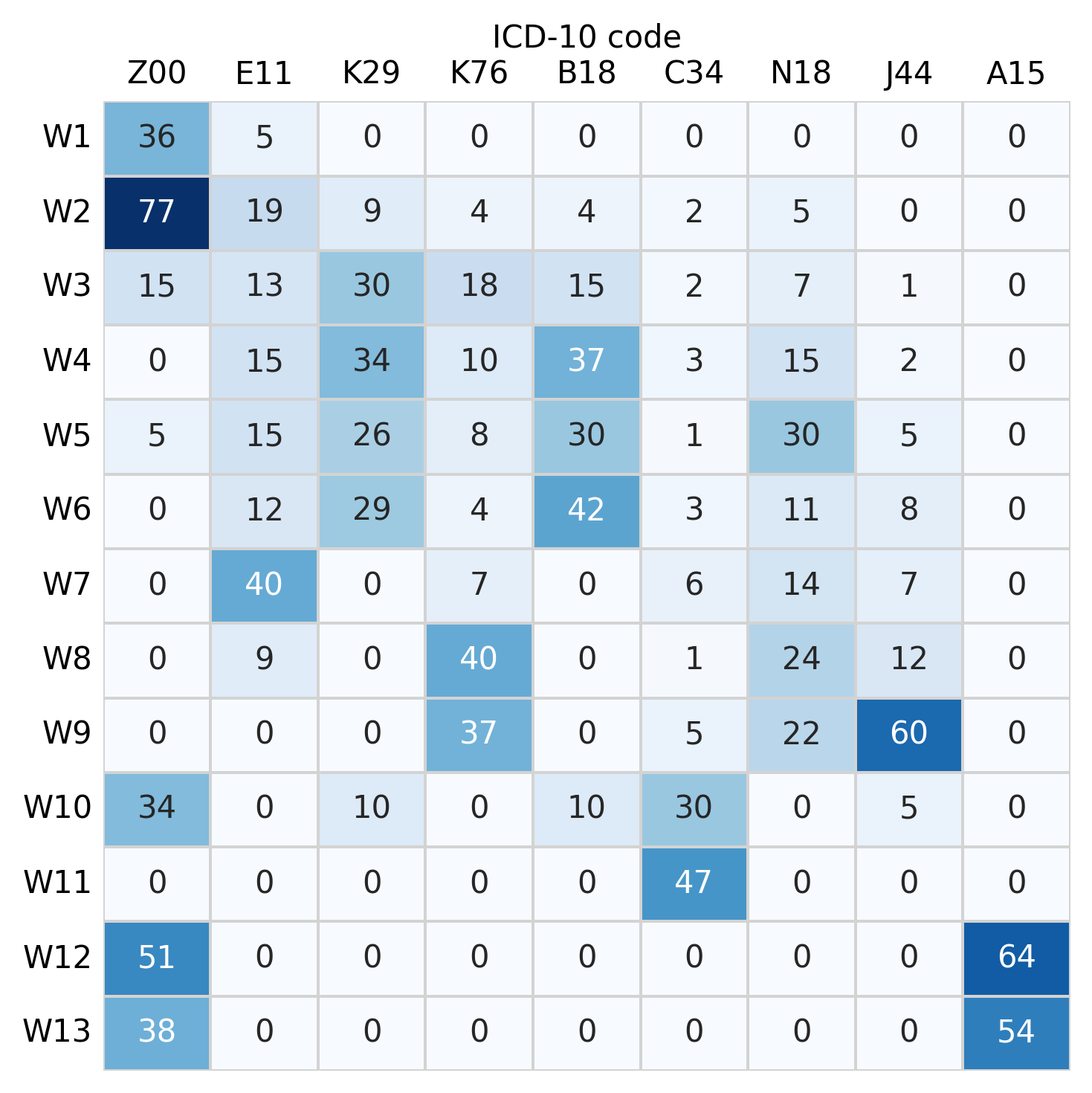}
    \end{center}
    \caption{Distribution of samples per week and diagnostic group. Each cell shows the number of samples. Weeks (W1-W13) are shown on the vertical axis, ICD-10 codes on the horizontal axis.} \label{fig:timeline}
\end{figure}

A key feature of S-OH is its explicit temporal annotation. The data were collected across consecutive weeks, with sample distribution shown in Figure~\ref{fig:timeline}. This temporal structure enables drift-aware evaluation: we partition weeks into training and test sets such that the classifier is evaluated on time periods not seen during training.

To minimize data leakage and simulate realistic deployment, we designed disease-specific temporal splits. For each medical condition, we selected different temporal segments for evaluation (Table~\ref{tab:splits}). For each disease, a subset of weeks was reserved for testing; the remaining weeks were used for training. Data from each patient is allocated entirely to either the test set or the training set.

\begin{table}
\centering
\normalsize
\caption{Train (empty) / test (+) split by ICD-10 codes across weeks. For each diagnostic group we use different temporal segments.} \label{tab:splits}
\setlength{\tabcolsep}{3.8pt}
\begin{tabular}{c *{9}{c}}
\toprule
\textbf{Week} & \textbf{Z00} & \textbf{E11} & \textbf{K29} & \textbf{K76} & \textbf{B18} & \textbf{C34} & \textbf{N18} & \textbf{J44} & \textbf{A15} \\
\midrule
W1  &   & + & + &   &   &   &   &   &   \\
W2  &   & + & + & + & + &   & + &   &   \\
W3  &   & + & + & + & + &   & + &   &   \\
W4  &   &   &   & + &   &   & + &   &   \\
W5  &   &   &   &   &   &   &   & + &   \\
W6  &   &   &   &   &   &   &   & + &   \\
W7  &   &   &   &   &   &   &   & + &   \\
W8  &   &   &   &   &   & + &   &   &   \\
W9  & + &   &   &   &   & + &   &   &   \\
W10 & + &   &   &   & + & + & + &   &   \\
W11 & + &   &   &   &   &   &   & + &   \\
W12 &   &   &   &   &   &   &   &   &   \\
W13 &   &   &   &   &   &   &   &   &  +\\
\bottomrule
\end{tabular}
\end{table}

We deliberately abstained from applying drift correction techniques in our core benchmarks. Instead, we preserve drift within the proposed temporal splits and scale each time series individually to reduce absolute baseline impact while preserving kinetic features. This design explicitly frames sensor drift robustness as a central, quantifiable challenge for the community.

\subsection{Data Format and Availability}
For each patient, the data are stored in a JSON structure containing demographic information (age, gender, ICD-10 diagnosis code), the eNose time series (17-channel multivariate signal, 895 seconds duration at 0.4 Hz), auxiliary sensor readings (temperature, pressure, humidity, CO$_2$ levels), and metadata (start/end timestamps, total duration).

The dataset and evaluation splits are publicly available at: \url{https://huggingface.co/datasets/ivanpodd/S-OH}.

\section{Experimental Setup}
\subsection{Evaluation Protocol}
All experiments follow the temporal train/test splits described in Section~\ref{Temporal_Splits}. For each disease, we train on samples from weeks designated as training and evaluate on samples from the held-out test weeks (Table~\ref{tab:splits}). This design ensures that training and test samples are temporally separated, simulating a realistic deployment scenario where the model cannot learn from future data.

To account for variability in training data composition, we report the mean and standard deviation of ROC AUC over 16 random splits of the training data. For each split, we randomly reserve 15\% of training samples for validation and early stopping, independent of temporal structure, as this subset is used solely for model selection and does not affect the final evaluation on the held-out test weeks.

\subsection{Feature-Based Methods}
We extract features from eight sensor channels that exhibited consistent signal quality. These channels were selected a priori based on two criteria: (i) the physical design of the multi-electrode chip, where certain channels were engineered for higher stability; and (ii) pre-screening tests conducted by the manufacturer, which identified these eight channels as the most reliable across repeated measurements.

Hyperparameters for ensemble methods were selected based on recommended defaults from the respective libraries. We did not perform extensive tuning, as our goal is to establish baseline performance rather than optimize individual methods.

Four feature sets are evaluated:
\begin{itemize}
    \item \textbf{Stats\_5}: minimum, maximum, mean, median, and standard deviation of each time series.
    \item \textbf{LogFit\_2}: we fit the logistic growth curve:
    \begin{equation}
        f(t) = R_0+\frac{R_{max}-R_0}{1+(\frac{t_{50}}{t})^k}. \label{LoqFit2}
    \end{equation}
    to each sensor channel's time series using \texttt{curve\_fit} from \texttt{scipy.optimize} and extract the two kinetic parameters $t_{50}$ and $k$ as features.
    \item \textbf{LogFit\_4}: all four parameters $R_{max}$, $R_0$, $t_{50}$, $k$ from Equation \ref{LoqFit2}.
    \item \textbf{Polynomial}: polynomial features extracted after wavelet smoothing and min-max normalization.
\end{itemize}

Before feature extraction, each time series is temporally clipped using experiment-defined \texttt{startTimeGases} and \texttt{endTimeGases} parameters, smoothed with a median filter to counter sensor degradation, and normalized via min-max scaling applied independently to each channel.

We train four ensemble methods on these feature sets. For \textbf{CatBoost}~\cite{prokhorenkova2018catboost}, we use 1500 iterations, a learning rate of 0.05, a maximum tree depth of 10, and L2 leaf regularization of 8. For \textbf{XGBoost}~\cite{chen2016xgboost}, we use 5000 iterations, a learning rate of 0.07, a maximum depth of 8, and an L2 penalty (reg\_lambda) of 2. For \textbf{HistGradientBoost}, we use 2000 iterations, a learning rate of 0.07, a maximum depth of 8, and L2 regularization of 0.2. For \textbf{RandomForest}, we use 200 estimators and the Gini impurity criterion.

\subsection{Convolution-Based Methods}
The key idea behind this approach is to treat time series data as images~\cite{hatami2018classification}. While 1D convolutions are computationally efficient, they would require learning temporal features from scratch. Our 2D approach allows transfer learning from ImageNet-pretrained models, which has been shown to improve performance on small medical datasets~\cite{matsoukas2022makes}.

\textbf{Preprocessing pipeline.} Each time series undergoes wavelet smoothing to reduce high-frequency noise while preserving signal shape, followed by min-max normalization applied independently to each channel to reduce absolute baseline variability. Next, we extract polynomial features (interaction terms and polynomial expansions) to capture non-linear dynamics between channels. Finally, the result is reshaped into a floating-point image-like matrix (width = time steps, height = features) suitable for convolutional architectures.

\textbf{Architectures.} We evaluate two convolutional models:
\begin{itemize}
    \item \textbf{CNN}: A shallow network with several convolutional and fully connected layers, trained from scratch on the polynomial feature matrices.
    \item \textbf{ResNet18}~\cite{he2016deep}: Pre-trained on ImageNet and fine-tuned for binary classification using the same polynomial feature input.
\end{itemize}

Both models achieve inference time under 1 second on CPU, making them suitable for embedded deployment on the eNose device itself.

\subsection{Graph-Based Method}
Given the eNose sensor principle that channel-to-channel ratios differ between samples of different origin, we hypothesize that a graph architecture could help leverage this distinct feature.

\textbf{Graph construction.} We build a complete graph with 8 nodes, each representing one of the eight stable sensor channels. For each node (channel $s_i$), we compute a 9-dimensional feature vector:
\begin{equation}
\begin{split}
x_i = \{min(s_i), max(s_i), mean(s_i), \\ std(s_i), k, t_{50}, a_0, a_1, a_2\},
\end{split}
\end{equation}
where $k$ and $t_{50}$ are kinetic parameters from Eq.~(\ref{LoqFit2}), and $a_0$, $a_1$, $a_2$ are Chebyshev polynomial coefficients approximating the temporal signal:
\begin{equation}
f(t) = a_0+a_1T_1(t)+a_2T_2(t),
\end{equation}
with $T_1$, $T_2$ denoting first- and second-order Chebyshev polynomials.

For each edge between nodes $i$ and $j$, we compute a 3-dimensional edge feature vector:
\begin{equation}
\begin{split}
x_{ij} = \{|min(s_i)-min(s_j)|, |max(s_i)-max(s_j)|, \\std(s_i^2)/std(s_j^2)\}.
\end{split}
\end{equation}

Before feature extraction, all time series are normalized independently to preserve channel-to-channel ratios while eliminating baseline-mediated drift. Each continuous value is discretized into 32 bins.

\textbf{Model.} We train the \textbf{Graphormer} model~\cite{ying2021transformers} for binary classification. Graphormer combines graph neural networks with transformer attention mechanisms, making it well-suited for complete graphs with rich edge features.

\subsection{Time Series Classification Methods}
As the data collected from the eNose device constitutes a multivariate time series, we apply three advanced time series models directly to raw signals (without manual feature engineering):
\begin{itemize}
    \item \textbf{LSTM}~\cite{karim2017lstm}: A 3-layer Long Short-Term Memory network with hidden state size of 128. LSTMs are well-suited for sequential data and can capture long-range temporal dependencies in breath signals.
    \item \textbf{InceptionTime}~\cite{ismail2020inceptiontime}: A deep ensemble of Inception-based architectures specifically designed for time series classification. Trained for 70 epochs with default hyperparameters. We report the performance of the full ensemble.
    \item \textbf{TS2Vec}~\cite{yue2022ts2vec}: An unsupervised representation learning framework for time series. We pre-train TS2Vec (learning rate 0.001, batch size 64, 40 epochs) to obtain universal representations of breath samples, then train a logistic regression classifier on top of the learned representations.
\end{itemize}

\subsection{Demographic Analysis}
A critical consideration in medical machine learning is the influence of demographic variables, which can be powerful predictors in themselves~\cite{rajkomar2018scalable}. To rigorously quantify the influence of demographics versus the breath signal, we perform three analyses:
\begin{enumerate}
    \item \textbf{Demographics-only baseline.} Train CatBoost models using only age, gender, and their combination (age+gender) to establish a lower bound of what can be predicted without breath data.
    \item \textbf{Gender-stratified evaluation.} Split the cohort into male (n=457) and female (n=567) subgroups and compute ROC AUC separately for each group.
    \item \textbf{Age-stratified evaluation.} Split the cohort at the median age (55 years) and evaluate performance on the older ($>$55) subgroup.
\end{enumerate}

This analysis reveals which diseases have strong demographic risk profiles versus those where breath signals provide independent diagnostic value.

\subsection{Drift Quantification}
To measure the severity of sensor drift--a known but rarely quantified failure mode in chemical sensing~\cite{bosch2022electronic}--we train a CatBoost model on the full dataset to predict the measurement day (from 1 to 71) using only raw sensor data, reporting the median absolute error (MAE) between predicted and true day.

The model achieves a median absolute error of 2-3 days, confirming that a strong temporal (non-stationary) signal is embedded in the sensor recordings. Notably, this prediction relied primarily on shifts in the absolute baseline rather than on dynamic signal features, suggesting that disease-related kinetic information may be partially separable from the drift component. This analysis justifies our use of temporal splits rather than random cross-validation: standard random splits would artificially mix temporally close samples, leading to overly optimistic performance estimates that do not generalize to real-world deployment.

\section{Results}
\subsection{Overall Classification Performance}
Table~\ref{tab:results_main} presents ROC AUC scores for all 10 methods across nine binary classification tasks. Several observations emerge.

\begin{table*}[h]
\centering
\caption{ROC AUC (mean{\scriptsize$\pm$std}) for binary disease classification. Best per disease in bold highlighting clinically promising results (AUC$>$0.65).} \label{tab:results_main}
\small
\begin{tabular}{lccccccccc}
\toprule
\textbf{Model + Features} & \textbf{Z00} & \textbf{E11} & \textbf{K29} & \textbf{K76} & \textbf{B18} & \textbf{C34} & \textbf{N18} & \textbf{J44} & \textbf{A15} \\
\midrule
\textit{LogisticRegression baseline} & 
\makecell{\textit{0.648} \\ \textit{{\scriptsize$\pm$0.038}}} & 
\makecell{\textit{0.524} \\ \textit{{\scriptsize$\pm$0.010}}} & 
\makecell{\textit{0.718} \\ \textit{{\scriptsize$\pm$0.010}}} & 
\makecell{\textit{0.302} \\ \textit{{\scriptsize$\pm$0.026}}} & 
\makecell{\textit{0.687} \\ \textit{{\scriptsize$\pm$0.024}}} & 
\makecell{\textit{0.457} \\ \textit{{\scriptsize$\pm$0.032}}} & 
\makecell{\textit{0.622} \\ \textit{{\scriptsize$\pm$0.015}}} & 
\makecell{\textit{0.533} \\ \textit{{\scriptsize$\pm$0.022}}} & 
\makecell{\textit{0.688} \\ \textit{{\scriptsize$\pm$0.044}}} \\
\multicolumn{9}{c}{\textit{Feature-based}} \\
CatBoost + LogFit\_2 & \makecell{0.465 \\ {\scriptsize$\pm$0.012}} & \makecell{0.516 \\ {\scriptsize$\pm$0.027}} & \makecell{0.448 \\ {\scriptsize$\pm$0.057}} & \makecell{0.612 \\ {\scriptsize$\pm$0.045}} & \makecell{0.686 \\ {\scriptsize$\pm$0.033}} & \makecell{0.503 \\ {\scriptsize$\pm$0.029}} & \makecell{0.598 \\ {\scriptsize$\pm$0.040}} & \makecell{0.448 \\ {\scriptsize$\pm$0.023}} & \makecell{0.552 \\ {\scriptsize$\pm$0.042}} \\
CatBoost + LogFit\_4 & \makecell{0.468 \\ {\scriptsize$\pm$0.006}} & \makecell{0.478 \\ {\scriptsize$\pm$0.020}} & \makecell{0.425 \\ {\scriptsize$\pm$0.033}} & \makecell{0.600 \\ {\scriptsize$\pm$0.061}} & \makecell{0.677 \\ {\scriptsize$\pm$0.028}} & \makecell{0.503 \\ {\scriptsize$\pm$0.028}} & \makecell{0.615 \\ {\scriptsize$\pm$0.028}} & \makecell{0.452 \\ {\scriptsize$\pm$0.018}} & \makecell{0.565 \\ {\scriptsize$\pm$0.038}} \\
CatBoost + Stats\_5 & \makecell{0.507 \\ {\scriptsize$\pm$0.016}} & \makecell{0.551 \\ {\scriptsize$\pm$0.028}} & \makecell{0.453 \\ {\scriptsize$\pm$0.032}} & \makecell{0.648 \\ {\scriptsize$\pm$0.027}} & \makecell{\textbf{0.698} \\ {\scriptsize$\pm$0.011}} & \makecell{0.530 \\ {\scriptsize$\pm$0.012}} & \makecell{0.541 \\ {\scriptsize$\pm$0.021}} & \makecell{0.435 \\ {\scriptsize$\pm$0.025}} & \makecell{0.574 \\ {\scriptsize$\pm$0.031}} \\
\midrule
XGBoost + LogFit\_2 & \makecell{0.455 \\ {\scriptsize$\pm$0.040}} & \makecell{0.508 \\ {\scriptsize$\pm$0.034}} & \makecell{0.458 \\ {\scriptsize$\pm$0.028}} & \makecell{0.545 \\ {\scriptsize$\pm$0.061}} & \makecell{0.605 \\ {\scriptsize$\pm$0.057}} & \makecell{0.549 \\ {\scriptsize$\pm$0.061}} & \makecell{0.507 \\ {\scriptsize$\pm$0.025}} & \makecell{0.462 \\ {\scriptsize$\pm$0.027}} & \makecell{0.566 \\ {\scriptsize$\pm$0.020}} \\
XGBoost + LogFit\_4 & \makecell{0.497 \\ {\scriptsize$\pm$0.008}} & \makecell{0.501 \\ {\scriptsize$\pm$0.022}} & \makecell{0.455 \\ {\scriptsize$\pm$0.049}} & \makecell{0.546 \\ {\scriptsize$\pm$0.036}} & \makecell{0.623 \\ {\scriptsize$\pm$0.063}} & \makecell{0.541 \\ {\scriptsize$\pm$0.061}} & \makecell{0.506 \\ {\scriptsize$\pm$0.040}} & \makecell{0.491 \\ {\scriptsize$\pm$0.030}} & \makecell{0.578 \\ {\scriptsize$\pm$0.040}} \\
XGBoost + Stats\_5 & \makecell{0.565 \\ {\scriptsize$\pm$0.013}} & \makecell{0.499 \\ {\scriptsize$\pm$0.044}} & \makecell{0.548 \\ {\scriptsize$\pm$0.057}} & \makecell{0.561 \\ {\scriptsize$\pm$0.057}} & \makecell{0.607 \\ {\scriptsize$\pm$0.040}} & \makecell{0.578 \\ {\scriptsize$\pm$0.035}} & \makecell{0.528 \\ {\scriptsize$\pm$0.048}} & \makecell{0.445 \\ {\scriptsize$\pm$0.046}} & \makecell{0.538 \\ {\scriptsize$\pm$0.029}} \\
\midrule
HGBoost + LogFit\_2 & \makecell{0.411 \\ {\scriptsize$\pm$0.038}} & \makecell{0.511 \\ {\scriptsize$\pm$0.045}} & \makecell{0.476 \\ {\scriptsize$\pm$0.045}} & \makecell{0.614 \\ {\scriptsize$\pm$0.042}} & \makecell{0.610 \\ {\scriptsize$\pm$0.044}} & \makecell{0.496 \\ {\scriptsize$\pm$0.036}} & \makecell{0.521 \\ {\scriptsize$\pm$0.030}} & \makecell{0.456 \\ {\scriptsize$\pm$0.037}} & \makecell{0.571 \\ {\scriptsize$\pm$0.035}} \\
HGBoost + LogFit\_4 & \makecell{0.440 \\ {\scriptsize$\pm$0.042}} & \makecell{0.495 \\ {\scriptsize$\pm$0.031}} & \makecell{0.470 \\ {\scriptsize$\pm$0.036}} & \makecell{0.563 \\ {\scriptsize$\pm$0.043}} & \makecell{0.676 \\ {\scriptsize$\pm$0.034}} & \makecell{0.463 \\ {\scriptsize$\pm$0.032}} & \makecell{0.515 \\ {\scriptsize$\pm$0.029}} & \makecell{0.481 \\ {\scriptsize$\pm$0.033}} & \makecell{0.600 \\ {\scriptsize$\pm$0.045}} \\
HGBoost + Stats\_5 & \makecell{0.510 \\ {\scriptsize$\pm$0.029}} & \makecell{0.526 \\ {\scriptsize$\pm$0.029}} & \makecell{0.515 \\ {\scriptsize$\pm$0.060}} & \makecell{0.615 \\ {\scriptsize$\pm$0.057}} & \makecell{0.651 \\ {\scriptsize$\pm$0.030}} & \makecell{0.580 \\ {\scriptsize$\pm$0.022}} & \makecell{0.561 \\ {\scriptsize$\pm$0.045}} & \makecell{0.429 \\ {\scriptsize$\pm$0.027}} & \makecell{0.550 \\ {\scriptsize$\pm$0.025}} \\
\midrule
RandomForest + LogFit\_2 & \makecell{0.479 \\ {\scriptsize$\pm$0.041}} & \makecell{0.491 \\ {\scriptsize$\pm$0.044}} & \makecell{0.471 \\ {\scriptsize$\pm$0.060}} & \makecell{0.588 \\ {\scriptsize$\pm$0.045}} & \makecell{0.627 \\ {\scriptsize$\pm$0.029}} & \makecell{0.519 \\ {\scriptsize$\pm$0.043}} & \makecell{0.530 \\ {\scriptsize$\pm$0.032}} & \makecell{0.475 \\ {\scriptsize$\pm$0.043}} & \makecell{0.531 \\ {\scriptsize$\pm$0.058}} \\
RandomForest + LogFit\_4 & \makecell{0.482 \\ {\scriptsize$\pm$0.041}} & \makecell{0.503 \\ {\scriptsize$\pm$0.041}} & \makecell{0.466 \\ {\scriptsize$\pm$0.046}} & \makecell{0.599 \\ {\scriptsize$\pm$0.051}} & \makecell{0.630 \\ {\scriptsize$\pm$0.061}} & \makecell{0.472 \\ {\scriptsize$\pm$0.047}} & \makecell{0.547 \\ {\scriptsize$\pm$0.055}} & \makecell{0.496 \\ {\scriptsize$\pm$0.031}} & \makecell{0.508 \\ {\scriptsize$\pm$0.041}} \\
RandomForest + Stats\_5 & \makecell{0.504 \\ {\scriptsize$\pm$0.029}} & \makecell{0.502 \\ {\scriptsize$\pm$0.048}} & \makecell{0.448 \\ {\scriptsize$\pm$0.052}} & \makecell{0.561 \\ {\scriptsize$\pm$0.080}} & \makecell{0.648 \\ {\scriptsize$\pm$0.036}} & \makecell{0.486 \\ {\scriptsize$\pm$0.027}} & \makecell{0.534 \\ {\scriptsize$\pm$0.044}} & \makecell{0.441 \\ {\scriptsize$\pm$0.035}} & \makecell{0.552 \\ {\scriptsize$\pm$0.050}} \\
\midrule
\multicolumn{9}{c}{\textit{Convolution-based}} \\
CNN + Polynomial & \makecell{0.600 \\ {\scriptsize$\pm$0.022}} & \makecell{0.468 \\ {\scriptsize$\pm$0.035}} & \makecell{0.545 \\ {\scriptsize$\pm$0.037}} & \makecell{0.403 \\ {\scriptsize$\pm$0.023}} & \makecell{0.511 \\ {\scriptsize$\pm$0.015}} & \makecell{0.700 \\ {\scriptsize$\pm$0.028}} & \makecell{0.636 \\ {\scriptsize$\pm$0.055}} & \makecell{0.470 \\ {\scriptsize$\pm$0.027}} & \makecell{0.567 \\ {\scriptsize$\pm$0.020}} \\
ResNet18 + Polynomial & \makecell{0.590 \\ {\scriptsize$\pm$0.061}} & \makecell{0.517 \\ {\scriptsize$\pm$0.039}} & \makecell{0.516 \\ {\scriptsize$\pm$0.042}} & \makecell{0.505 \\ {\scriptsize$\pm$0.030}} & \makecell{0.575 \\ {\scriptsize$\pm$0.042}} & \makecell{0.612 \\ {\scriptsize$\pm$0.033}} & \makecell{0.630 \\ {\scriptsize$\pm$0.041}} & \makecell{0.464 \\ {\scriptsize$\pm$0.049}} & \makecell{0.548 \\ {\scriptsize$\pm$0.013}}\\
\midrule
\multicolumn{9}{c}{\textit{Graph-based}} \\
Graphormer & \makecell{0.578 \\ {\scriptsize$\pm$0.118}} & \makecell{0.554 \\ {\scriptsize$\pm$0.075}} & \makecell{0.513 \\ {\scriptsize$\pm$0.051}} & \makecell{0.502 \\ {\scriptsize$\pm$0.113}} & \makecell{0.462 \\ {\scriptsize$\pm$0.044}} & \makecell{0.630 \\ {\scriptsize$\pm$0.059}} & \makecell{0.517 \\ {\scriptsize$\pm$0.095}} & \makecell{\textbf{0.767} \\ {\scriptsize$\pm$0.061}} & \makecell{0.477 \\ {\scriptsize$\pm$0.054}} \\
\midrule
\multicolumn{9}{c}{\textit{Time series}} \\
LSTM & \makecell{0.484 \\ {\scriptsize$\pm$0.108}} & \makecell{0.493 \\ {\scriptsize$\pm$0.030}} & \makecell{\textbf{0.658} \\ {\scriptsize$\pm$0.064}} & \makecell{\textbf{0.686} \\ {\scriptsize$\pm$0.017}} & \makecell{0.613 \\ {\scriptsize$\pm$0.010}} & \makecell{0.691 \\ {\scriptsize$\pm$0.033}} & \makecell{\textbf{0.658} \\ {\scriptsize$\pm$0.032}} & \makecell{0.626 \\ {\scriptsize$\pm$0.031}} & \makecell{0.476 \\ {\scriptsize$\pm$0.048}} \\
InceptionTime & \makecell{0.496 \\ {\scriptsize$\pm$0.121}} & \makecell{0.492 \\ {\scriptsize$\pm$0.053}} & \makecell{0.495 \\ {\scriptsize$\pm$0.074}} & \makecell{0.408 \\ {\scriptsize$\pm$0.048}} & \makecell{0.548 \\ {\scriptsize$\pm$0.060}} & \makecell{\textbf{0.746} \\ {\scriptsize$\pm$0.030}} & \makecell{0.608 \\ {\scriptsize$\pm$0.065}} & \makecell{0.461 \\ {\scriptsize$\pm$0.057}} & \makecell{0.489 \\ {\scriptsize$\pm$0.041}} \\
TS2Vec & \makecell{0.480 \\ {\scriptsize$\pm$0.029}} & \makecell{0.466 \\ {\scriptsize$\pm$0.032}} & \makecell{0.453 \\ {\scriptsize$\pm$0.092}} & \makecell{0.538 \\ {\scriptsize$\pm$0.034}} & \makecell{0.644 \\ {\scriptsize$\pm$0.017}} & \makecell{0.579 \\ {\scriptsize$\pm$0.030}} & \makecell{0.654 \\ {\scriptsize$\pm$0.028}} & \makecell{0.470 \\ {\scriptsize$\pm$0.049}} & \makecell{0.474 \\ {\scriptsize$\pm$0.028}} \\
\bottomrule
\end{tabular}
\end{table*}

First, the best results are achieved for lung cancer (C34), hepatitis (B18), and COPD (J44), with several methods surpassing or approaching AUC 0.70. InceptionTime achieves the highest score for lung cancer (AUC=0.746), Graphormer achieves the best for COPD (AUC=0.767), and CatBoost with Stats\_5 features achieves the best for hepatitis (AUC=0.698).

Second, performance varies dramatically across diseases. While lung cancer and COPD show promising results, diabetes (E11) and respiratory tuberculosis (A15) remain near chance (AUC in range 0.45-0.55) across all methods, suggesting that the breath signal for these conditions is either weak, confounded, or beyond the detection limits of the current sensor technology.

Third, among feature-based methods, Stats\_5 features consistently outperforms the kinetic feature sets (LogFit\_2, LogFit\_4) across most ensemble models, indicating that simple statistical summaries capture more discriminative information than fitted growth curve parameters for this dataset.

Fourth, to identify which methods generalize best across diseases, we rank methods by their average AUC across all nine conditions (Table~\ref{tab:ranking}). LSTM achieves the highest average rank (Avg AUC=0.598), followed by Graphormer (0.556) and ResNet18 (0.551). Notably, Graphormer exhibits one of the highest variance (Std AUC=0.095), performing exceptionally well on COPD (AUC=0.767) but near-random on other diseases, suggesting that the model's performance is highly disease-dependent and may not generalize well across the full diagnostic spectrum. LSTM, in contrast, shows the most consistent performance across diseases, ranking in the top three for 5 out of 9 conditions.

\begin{table}[h]
\centering
\caption{Method ranking by average ROC AUC across all diseases.} \label{tab:ranking}
\small
\begin{tabular}{l c c}
\toprule
\textbf{Method} & \textbf{Avg AUC} & \textbf{Std AUC} \\
\midrule
LSTM & 0.598 & 0.089 \\
Graphormer & 0.556 & 0.095 \\
ResNet18 + Polynomial & 0.551 & 0.055 \\
HGBoost + Stats\_5 & 0.549 & 0.065 \\
CatBoost + Stats\_5 & 0.549 & 0.084 \\
CNN + Polynomial & 0.544 & 0.093 \\
XGBoost + Stats\_5 & 0.541 & 0.047 \\
TS2Vec & 0.529 & 0.079 \\
InceptionTime & 0.527 & 0.099 \\
RandomForest + LogFit\_2 & 0.523 & 0.054 \\
\bottomrule
\end{tabular}
\end{table}

\subsection{Impact of Sensor Drift}
To quantify the effect of temporal drift on classification performance, we compare our temporal split evaluation against a idealized random split baseline. Table~\ref{tab:drift_impact} shows that all methods experience significant degradation when evaluated on temporally separated data. The average AUC drop across all diseases and methods is approximately 0.05-0.12.

\begin{table}[h]
\centering
\caption{Effect of Random vs Temporal split on ROC AUC (selected methods).}
\label{tab:drift_impact}
\small
\begin{tabular}{p{2.4cm} c c c}
\toprule
\textbf{Disease} & \textbf{Method} & \textbf{Random} & \textbf{Temporal} \\
\midrule
Lung cancer (C34) & InceptionTime & 0.812 & 0.746 \\
Hepatitis (B18) & CatBoost+Stats\_5 & 0.751 & 0.698 \\
COPD (J44) & Graphormer & 0.831 & 0.767 \\
Healthy (Z00) & CNN & 0.674 & 0.600 \\
\bottomrule
\end{tabular}
\end{table}

\subsection{Demographic Confounders}
Table~\ref{tab:demographics} presents results of demographic-only models and subgroup analysis. Several findings are notable.

\begin{table*}[h]
\centering
\caption{Demographic analysis: ROC AUC for CatBoost + Stats\_5.}
\label{tab:demographics}
\small
\begin{tabular}{lccccccccc}
\toprule
\textbf{Cohort} & \textbf{Z00} & \textbf{E11} & \textbf{K29} & \textbf{K76} & \textbf{B18} & \textbf{C34} & \textbf{N18} & \textbf{J44} & \textbf{A15} \\
\midrule
\multicolumn{9}{c}{\textit{Demographics-only models}} \\
Age only & \makecell{0.967 \\ {\scriptsize$\pm$0.004}} & \makecell{0.627 \\ {\scriptsize$\pm$0.011}} & \makecell{0.494 \\ {\scriptsize$\pm$0.020}} & \makecell{0.473 \\ {\scriptsize$\pm$0.031}} & \makecell{0.555 \\ {\scriptsize$\pm$0.012}} & \makecell{0.678 \\ {\scriptsize$\pm$0.009}} & \makecell{0.617 \\ {\scriptsize$\pm$0.011}} & \makecell{0.555 \\ {\scriptsize$\pm$0.017}} & \makecell{0.803 \\ {\scriptsize$\pm$0.010}} \\
Gender only & \makecell{0.570 \\ {\scriptsize$\pm$0.000}} & \makecell{0.520 \\ {\scriptsize$\pm$0.000}} & \makecell{0.499 \\ {\scriptsize$\pm$0.000}} & \makecell{0.497 \\ {\scriptsize$\pm$0.000}} & \makecell{0.607 \\ {\scriptsize$\pm$0.000}} & \makecell{0.607 \\ {\scriptsize$\pm$0.000}} & \makecell{0.500 \\ {\scriptsize$\pm$0.078}} & \makecell{0.616 \\ {\scriptsize$\pm$0.000}} & \makecell{0.712 \\ {\scriptsize$\pm$0.009}} \\
Age + gender & \makecell{0.935 \\ {\scriptsize$\pm$0.012}} & \makecell{0.651 \\ {\scriptsize$\pm$0.008}} & \makecell{0.575 \\ {\scriptsize$\pm$0.025}} & \makecell{0.447 \\ {\scriptsize$\pm$0.025}} & \makecell{0.583 \\ {\scriptsize$\pm$0.012}} & \makecell{0.729 \\ {\scriptsize$\pm$0.007}} & \makecell{0.632 \\ {\scriptsize$\pm$0.013}} & \makecell{0.651 \\ {\scriptsize$\pm$0.008}} & \makecell{0.923 \\ {\scriptsize$\pm$0.012}} \\
\midrule
\multicolumn{9}{c}{\textit{Subgroup analysis (CatBoost + Stats\_5)}} \\
Male & \makecell{0.597 \\ {\scriptsize$\pm$0.014}} & \makecell{0.474 \\ {\scriptsize$\pm$0.047}} & \makecell{0.380 \\ {\scriptsize$\pm$0.062}} & \makecell{0.484 \\ {\scriptsize$\pm$0.055}} & \makecell{0.552 \\ {\scriptsize$\pm$0.046}} & \makecell{0.477 \\ {\scriptsize$\pm$0.055}} & \makecell{0.562 \\ {\scriptsize$\pm$0.049}} & \makecell{0.462 \\ {\scriptsize$\pm$0.022}} & \makecell{0.467 \\ {\scriptsize$\pm$0.042}} \\
Female & \makecell{0.382 \\ {\scriptsize$\pm$0.016}} & \makecell{0.459 \\ {\scriptsize$\pm$0.032}} & \makecell{0.422 \\ {\scriptsize$\pm$0.046}} & \makecell{0.653 \\ {\scriptsize$\pm$0.035}} & \makecell{0.763 \\ {\scriptsize$\pm$0.039}} & \makecell{0.477 \\ {\scriptsize$\pm$0.052}} & \makecell{0.653 \\ {\scriptsize$\pm$0.028}} & \makecell{0.615 \\ {\scriptsize$\pm$0.038}} & \makecell{0.606 \\ {\scriptsize$\pm$0.036}} \\
Age $>$55 & - & \makecell{0.520 \\ {\scriptsize$\pm$0.040}} & \makecell{0.451 \\ {\scriptsize$\pm$0.052}} & \makecell{0.567 \\ {\scriptsize$\pm$0.056}} & \makecell{0.684 \\ {\scriptsize$\pm$0.045}} & \makecell{0.730 \\ {\scriptsize$\pm$0.021}} & \makecell{0.549 \\ {\scriptsize$\pm$0.041}} & \makecell{0.464 \\ {\scriptsize$\pm$0.027}} & \makecell{0.598 \\ {\scriptsize$\pm$0.029}} \\
\bottomrule
\end{tabular}
\end{table*}

\textbf{Demographics-only baselines.} It is interesting to note that the exceptionally high age-only AUC for healthy controls (0.967) reflects the age distribution of the Z00 group, which contains predominantly younger individuals compared to most disease groups. For lung cancer (C34) and respiratory tuberculosis (A15), age alone achieves AUC of 0.678 and 0.803 respectively, rivaling or even exceeding some sensor-based models. This reflects the well-known age dependence of these diseases, as both lung cancer incidence and tuberculosis risk increase significantly in older populations \cite{li2025global,yayan2026tuberculosis}. In contrast, for liver disease (K76) and gastritis and duodenitis (K29), demographic-only models perform poorly (AUC$\leq$0.5), yet sensor-based models achieve AUC$>$0.65, indicating that breath signals provide substantial independent diagnostic value.

\textbf{Subgroup analysis.} Performance varies across demographic subgroups. For hepatitis (B18), classification is substantially better in women (AUC=0.763) than in men (AUC=0.552). For lung cancer (C34), the older cohort (age $>$55) shows substantially higher AUC (0.730) compared to the overall model (0.530), reflecting the strong age dependence of this disease. These differences suggest that model performance may be influenced by underlying differences in disease manifestation or breath VOC profiles across demographic groups.

\section{Discussion}
Our benchmark reveals four actionable insights for researchers and practitioners developing eNose-based disease screening systems.

\subsection{Performance Ceiling}
Interestingly, the simple Logistic Regression baseline achieves competitive performance on several diseases (e.g., AUC=0.718 for K29, $>$0.68 for B18 and A15). Notably, some advanced methods fail to consistently outperform this linear baseline, suggesting that careful evaluation against simple models is essential before claiming algorithmic progress in clinical breathomics. The best methods achieve ROC AUC of 0.74-0.76 for lung cancer (C34) and COPD (J44), demonstrating that breath signals contain clinically relevant information for these conditions. However, performance varies dramatically across diseases. While lung cancer and COPD show promising results, diabetes (E11) and respiratory tuberculosis (A15) remain near chance across all 10 methods.

This variation likely reflects fundamental differences in disease biology. Lung cancer and COPD are associated with distinct metabolic alterations that produce strong VOC signatures in exhaled breath~\cite{li2023research}. In contrast, while diabetes--despite being a metabolic disease--may be confounded by the fact that many diabetic patients in our cohort were undergoing treatment, potentially normalizing their breath profiles at the time of sampling.

For diseases where breath signals are weak, alternative screening approaches or multi-modal models (combining eNose with demographic or clinical variables) may be necessary. The high performance on lung cancer and COPD, however, suggests that eNose technology is already viable as a triage tool for these conditions in high-risk populations.

Graphormer's exceptional performance on COPD (AUC=0.767) and high variance across diseases suggests that graph-based methods may require larger datasets or stronger regularization to generalize across the full disease spectrum. The near-chance performance for tuberculosis (A15), in contrast to prior literature~\cite{bruins2013diagnosis}, may reflect that our A15 cohort was collected at a dedicated treatment center, potentially normalizing breath VOC profiles at the time of sampling.

\subsection{Drift Is Not Harmless: A Critical Deployment Barrier}
All methods degrade significantly when tested on temporally separated data (Table~\ref{tab:drift_impact}). The average AUC drop of 0.05-0.12 confirms that standard random cross-validation (still common in the literature) substantially overestimates real-world performance.

Researchers should abandon random cross-validation for eNose studies. Temporal validation is the minimum standard for realistic performance estimation. While we do not benchmark active drift-adaptation methods here, our temporal splits and baselines provide a foundation for evaluating such techniques (e.g., domain-adversarial training~\cite{ganin2016domain} or style transfer) in future work.

\subsection{No Single Winner: Method Selection Depends on Priorities}
Our benchmark reveals a trade-off between consistency and peak performance. LSTM shows the most consistent performance across diseases, ranking in the top three for 5 out of 9 conditions (Table~\ref{tab:ranking}). This suggests that the recurrent architecture, which explicitly models temporal dependencies, is well-suited to the sequential nature of breath signals.

In contrast, Graphormer achieves the best result for COPD (AUC=0.767) and InceptionTime for lung cancer (AUC=0.746), but both perform poorly on other diseases. This specialization may reflect the sensitivity of these architectures to disease-specific temporal patterns.

Feature-based methods (particularly CatBoost with Stats\_5) remain competitive, achieving the best result for hepatitis (AUC=0.698) and ranking fourth overall by average AUC. Their simplicity, interpretability, and low computational cost make them attractive for clinical deployment, especially in resource-constrained settings.

Method selection should be guided by deployment priorities: LSTM for consistency across diseases, Graphormer or InceptionTime for peak performance on a specific target, CatBoost with Stats\_5 features for low cost and interpretability.

\subsection{Demographic Confounders: Signal vs Bias}
Demographic variables can be powerful predictors in medical ML~\cite{rajkomar2018scalable}. Our analysis reveals a nuanced picture.

For lung cancer (C34) and tuberculosis (A15), age alone rivals sensor-based models (AUC=0.678 and 0.803, respectively), raising the question of whether models learn biology or simply demographic trends. In contrast, for liver disease and gastritis (K76 and K29), demographic-only models perform poorly (AUC about 0.5) while sensor-based models exceed AUC$>$0.65, providing strong evidence that breath signals carry independent diagnostic value.

Researchers should always report demographic-stratified performance and compare sensor-based models against demographic-only baselines. A model that outperforms age and gender on diseases with weak demographic risk profiles provides stronger evidence of genuine VOC signal than one that merely reflects known demographic trends.

\subsection{Limitations and Future Directions}
Several limitations should be acknowledged. While our dataset includes patients from two clinical sites, both used the same eNose device. Cross-device and cross-batch validation are needed to assess generalization to different sensor hardware and manufacturing batches. Additionally, we did not apply drift correction techniques; our benchmark establishes baseline performance, providing a foundation for future work on drift-robust methods.

Future work should focus on three directions: (i) drift-robust learning (domain adaptation, style transfer, self-supervised pretraining) to reduce the 0.05-0.12 AUC drop; (ii) multi-disease models exploiting correlations among conditions; and (iii) prospective validation on new patient cohorts, which our temporal splits approximate but cannot fully replace.

\section{Conclusion}
We introduced Breath-Dx, the first comprehensive benchmark of time series classification methods for breath-based disease screening using electronic nose technology. Our work makes three primary contributions.

First, we established a reproducible evaluation framework for clinical eNose data, quantifying for the first time the true impact of sensor drift: a 0.05-0.12 AUC drop under temporal validation, demonstrating that random cross-validation is inappropriate for such studies.

Second, we benchmarked 10 time series methods on clinical breath data, revealing that no single method dominates all scenarios. Recurrent architectures offer the best cross-disease stability, while graph-based and specialized time series methods achieve the highest peak performance (AUC$>$0.74) on specific diseases (lung cancer and COPD) at the cost of cross-disease generalization.

Third, we derived actionable insights for clinical deployment. A simple Logistic Regression baseline outperforms several advanced methods on certain diseases, indicating that not all algorithmic complexity translates into clinical benefit. Breath signals provide independent diagnostic value beyond demographics for liver disease and gastritis (sensor-based AUC$>$0.65 vs demographics near chance), while for lung cancer and tuberculosis, age alone achieves AUC up to 0.803, rivaling or exceeding sensor-based models. These findings highlight the need to explicitly account for demographic confounders and to always compare against simple baselines in eNose screening studies.

To enable reproducible research and accelerate progress in clinical breathomics, we release the full Breath-Dx benchmark, including all temporal splits, evaluation protocols, and code, at \url{https://github.com/ivanpodd/Breath-Dx}.

We hope Breath-Dx serves as a foundational resource for the community, lowering the barrier to entry for researchers developing robust, generalizable machine learning for non-invasive disease screening.

\section*{Acknowledgment}
The authors thank Fedor Fedorov and the Skoltech Laboratory of Nanomaterials (led by Professor Albert Nasibulin) for their invaluable contributions, particularly Valerii Zaitsev (crystal testing), Maxim Fomenko (chip printing), and Ilia Tronev (nanomaterial synthesis). We also thank Timur Asyaev and Stanislav Pesnya of Sber's Internet of Things Laboratory for the technical implementation of the eNose device. Finally, we acknowledge the teams at the Moscow Regional Research and Clinical Institute and the Central Research Institute of Tuberculosis for assistance with patient data collection.

\bibliographystyle{IEEEtran}
\bibliography{bibfile}

\end{document}